\documentclass{article}

\usepackage{PRIMEarxiv}
\usepackage{amssymb}

\usepackage[utf8]{inputenc} 
\usepackage[T1]{fontenc}    
\usepackage{hyperref}       
\usepackage{url}            
\usepackage{booktabs}       
\usepackage{amsfonts}       
\usepackage{nicefrac}       
\usepackage{microtype}      
\usepackage{lipsum}
\usepackage{fancyhdr}       
\usepackage{graphicx}       

\usepackage{graphicx, subcaption}
\usepackage{algorithm} 
\usepackage{multirow}
\usepackage{indentfirst}
\usepackage[normalem]{ulem}
\usepackage[x11names]{xcolor}
\usepackage[euler]{textgreek}
\usepackage{amsmath}
\usepackage{amsthm}
\usepackage{mathtools, nccmath}
\usepackage{indentfirst}
\usepackage[normalem]{ulem}
\usepackage[font = {footnotesize}]{caption}
\usepackage{multirow}
\usepackage{setspace} 

\newcommand{\specialcell}[2][c]{%
  \begin{tabular}[#1]{@{}c@{}}#2\end{tabular}}

\newtheoremstyle{typex-definition} 
  {3pt}       
  {3pt}       
  {\itshape}  
  {}          
  {\bfseries\itshape}           
  {\textcolor{black}{.}}  
  { }         
  {%
   \thmname{#1}
   \thmnumber{ \textcolor{black}{#2}}
   \thmnote{ {\normalfont\itshape(#3)}}%
 }

\theoremstyle{typex-definition}
\newtheorem{definition}{\textcolor{black}{Definition}}[section]

\AtBeginDocument{%
  \providecommand\BibTeX{{%
    \normalfont B\kern-0.5em{\scshape i\kern-0.25em b}\kern-0.8em\TeX}}}

\title{Fairguard: Harness Logic-based Fairness Rules in Smart Cities
}

\author{
  Yiqi Zhao \\
  Vanderbilt University \\
\texttt{yiqi.zhao@vanderbilt.edu} \\
   \And
  Ziyan An \\
  Vanderbilt University\\
  \texttt{ziyan.an@vanderbilt.edu} \\
  \AND
  Xuqing Gao \\
  Vanderbilt University\\
  \texttt{xuqing.gao@vanderbilt.edu} \\
  \And
  Ayan Mukhopadhyay\\
  Vanderbilt University\\
  \texttt{ayan.mukhopadhyay@vanderbilt.edu} \\
  \And
  Meiyi Ma \\
  Vanderbilt University \\
  \texttt{meiyi.ma@vanderbilt.edu} \\
}

\begin{document}
\maketitle

\begin{abstract}
Smart cities operate on computational predictive frameworks that collect, aggregate, and utilize data from large-scale sensor networks. However, these frameworks are prone to multiple sources of data and algorithmic bias, which often lead to unfair prediction results. In this work, we first demonstrate that bias persists at a micro-level both temporally and spatially by studying real city data from Chattanooga, TN. To alleviate the issue of such bias, we introduce \textit{Fairguard}, a micro-level temporal logic-based approach for fair smart city policy adjustment and generation in complex temporal-spatial domains. The Fairguard framework consists of two phases: first, we develop a static generator that is able to reduce data bias based on temporal logic conditions by minimizing correlations between selected attributes. Then, to ensure fairness in predictive algorithms, we design a dynamic component to regulate prediction results and generate future fair predictions by harnessing logic rules. Evaluations show that logic-enabled static Fairguard can effectively reduce the biased correlations while dynamic Fairguard can guarantee fairness on protected groups at run-time with minimal impact on overall performance. 
\end{abstract}


\section{Introduction}
\label{sec:introduction}

Smart cities are large-scale computational systems with connected sensors collecting real-time data streams for more efficient, diverse, and reliable city operations. 
For example, intelligent transportation services have been designed as a smart city component to provide faster and more resilient traffic mobility~\cite{menouar2017uav}. 
At the center of smart city operations is data-driven decision-making. Such decisions are typically performed as part of a \textit{predict-then-optimize} loop, where a random variable of interest, representing the demand of a resource, is predicted and then subsequently used to allocate the resource to the population~\cite{elmachtoub2022smart}. 
For example, fields like public transit, ride-sharing services, emergency response, and electric vehicle infrastructure rely on forecasting models that estimate the likelihood of risk (or reward) of various actions to improve decision-making~\cite{pettet2021hierarchical,sivagnanam2022offline,alonso2017demand}. Consider on-demand transit services or shared micro-transit as examples; vehicles and bike stations are positioned dynamically based on predicted ridership demand~\cite{davis2016multi}. While the availability of forecasting models is crucial to the design of principled decision-making pipelines~\cite{elmachtoub2022smart}, the use of data-driven forecasting models can introduce (or reinforce) existing socio-economic disparities by encoding existing biases in the statistical models~\cite{corbett2018measure}. Perhaps more worryingly, predictive models are often integrated as black-box components in these decision-making pipelines, hampering interpretability and investigation of their consequences in practice~\cite{luckey2020artificial}. The effects of such biases have been widely studied~\cite{corbett2018measure,kallus2018residual} and algorithmic approaches have been designed to mitigate the unfairness that data-driven learning can inadvertently propagate. 


Fairness in data-driven learning has been studied extensively~\cite{corbett2017algorithmic} (it has also been studied in the context of decision-making, albeit to a significantly smaller extent). In order to evaluate and mitigate the effects of \textit{unfairness}, it is imperative that we first define fairness. While several formal definitions exist, the central theme revolves around predictive outcomes being \textit{equal} among groups defined by protected attributes or features (e.g., race or gender). Admittedly, our coverage of such definitions is rather loose (and outside the scope of this paper); we refer interested readers to  Corbett-Davies et al.~\cite{corbett2017algorithmic}, who present an excellent overview of different notions of fairness. Instead, we focus on how such notions are implemented as part of smart city operations. Specifically, we point out that while the problem of fairness is encoded as a \textit{static} component in decision-making pipelines (where static refers to fixed with respect to time or space), in practice, biases in data evolve dynamically.

\begin{figure}
    \centering
    \captionsetup{justification=centering}
    \includegraphics[width=.75\textwidth]{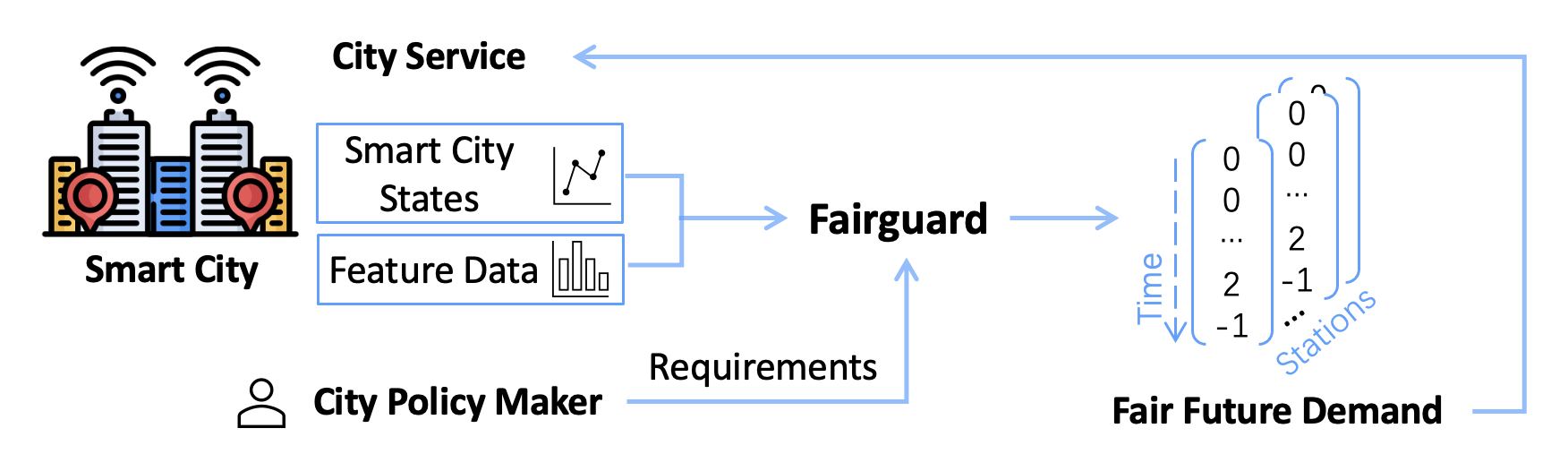}
    \caption{Fairguard overview (feature data - demographic or socioeconomic feature values \\ associated with studied city regions, e.g., household mean income)}
    \label{fig:cityOverview}
\end{figure}

Two popular approaches for mitigating the effects of data-driven forecasting are identifying the existing biases in datasets~\cite{chouldechova2020snapshot} and enforcing fairness by trading off model performance~\cite{roh2021sample,8970908,elazar2018adversarial}.
For example, bias in training data can be removed during pre-processing by modifying the data, and standard ways to enforce algorithmic fairness are to find a better learning objective or use fairness goals as a regularizer~\cite{corbett2017algorithmic, goel2017metric, yan2020fairness}. 
However, such methods consider bias (and subsequently fairness) as fundamentally static, i.e., bias in data and the objectives for achieving fair predictions is fixed (e.g., through summary statistics or expected values of random variables over space and time) and rarely updated. In this paper, we study the problem of shared micro-transit services, a classic human-in-the-loop cyber-physical system. We show that bias in data pertaining to the usage of shared bicycles (which is then used to predict the demand of bicycles) exhibits biases in correlation with several attributes in manners that change over time and space.
A critical challenge in designing such data-driven smart city infrastructure is constant monitoring of biased data and downstream predictions. We hypothesize that dynamic and simultaneous updates are required on both the data and the learning algorithm (often at micro-levels); such adjustments can take into account temporal and spatial variations to ensure fair outcomes.

To tackle this challenge, we introduce \textit{Fairguard}, a temporal logic-based micro-level framework for smart city policy generation and adjustment. 
More specifically, Fairguard is a two-phase model-agnostic framework that works with any gradient-based predictive models such as recurrent neural networks, convolutional neural networks, and graph neural networks.
To account for biased correlations in the training data, real-time smart city states collected from sensors are fed into Fairguard, which also takes in a set of pre-defined fairness constraints specified in Signal Temporal Logic (STL)~\cite{ma2021novel, maler2004monitoring}. Finally, to mitigate undesired algorithmic bias in predictive models in real time, the dynamic component of Fairguard performs fair rule distillation under micro-level logic specifications to correct model predictions. Specifically, we make the following contributions:
\begin{enumerate}
  \item We present four comprehensive case studies that highlight the problem of fairness in smart city services by focusing on data bias in shared micromobility in Chattanooga, TN from 2012 to 2021 and identify key challenges to developing fair smart city services.
  Specifically, we consider the shared bicycle service operated by the Chattanooga Regional Transportation Authority and analyze the bike demand by census tract under the influence of several economic and demographic factors. 
  \item We design \textit{Fairguard}, a lightweight, model-agnostic two-stage framework for deep learning algorithms, to ensure that bias is mitigated from both the training data and the predictive algorithm. The first stage features a logic-based flexible data adjustment process, while the second stage features a real-time monitoring structure. 
  \item We use Fairguard to adjust training data for a smart city service empirically and adjust model predictions dynamically by recommending fair city states. At the same time, we allow the adjustments to be more flexible and the recommended city states to deviate the least from the original policy while focusing on guarding fairness.
  \item We integrate our framework into state-of-the-art deep learning algorithms and evaluate the dynamic component of Fairguard using bicycle transit data in Chattanooga, TN. The framework is able to reallocate resources to ensure the demands of protected regions are met while taking into consideration available city resources.
\end{enumerate}

The rest of this paper is organized as follows. In Section~\ref{sec:moti}, we describe a motivating study of smart city fairness issues, using the bicycle transit system as an example. In Section~\ref{sec:overview}, we provide an overview of the Fairguard framework. In Section~\ref{sec:methods}, we present comprehensive technical details of Fairguard. Next, in Section~\ref{sec:evaluation}, we empirically show how Fairguard can help improve fairness. In Section~\ref{sec:relatedwork}, we discuss previous and related work. Lastly, in Section~\ref{sec:conclusion}, we conclude this study. 

\begin{figure*}[t]
\minipage{0.33\textwidth}
  \includegraphics[width=\linewidth]{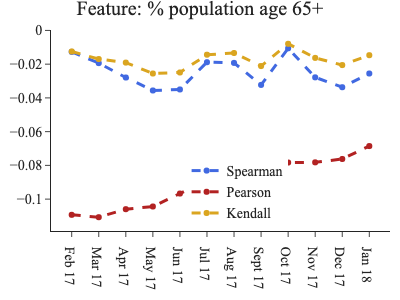}
\endminipage\hfill
\minipage{0.33\textwidth}
  \includegraphics[width=\linewidth]{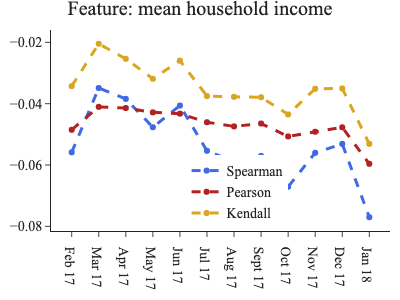}
\endminipage\hfill
\minipage{0.33\textwidth}%
  \includegraphics[width=\linewidth]{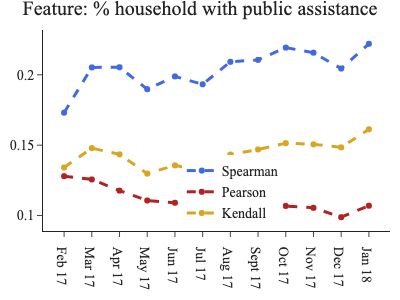}
\endminipage
\caption{Spatial correlation coefficient for bike demands with respect to three features (percentage of population over the age of 65, mean household income, and percentage of household with public assistance).}\label{fig:spatial}
\end{figure*}

\begin{figure*}[t]
\minipage{0.33\textwidth}
  \includegraphics[width=\linewidth]{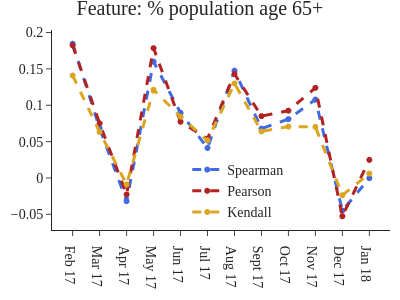}
\endminipage\hfill
\minipage{0.33\textwidth}
  \includegraphics[width=\linewidth]{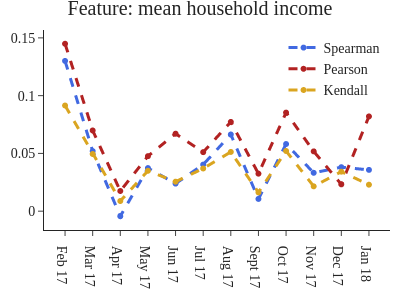}
\endminipage\hfill
\minipage{0.33\textwidth}%
  \includegraphics[width=\linewidth]{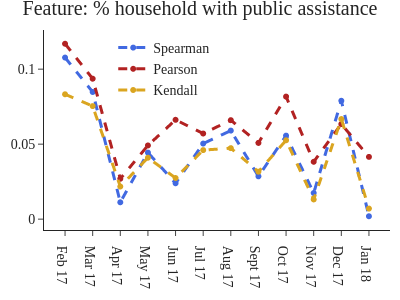}
\endminipage
\caption{Temporal correlation coefficient for bike demands with respect to three features (percentage of population over the age of 65, mean household income, and percentage of household with public assistance).}\label{fig:temporal}
\end{figure*}

\section{Motivating Study}
\label{sec:moti}

In this section, we present a motivating study on shared micro-transit data from Chattanooga that shows the dynamic changes of correlations between the service demand and other relevant city neighborhood characteristics (that may influence the demand). We use historical ridership as a proxy for demand for the ease of exposition. We collect data from 42 city-wide bicycle stations in Chattanooga, TN from July 2012 to July 2021 
for a comprehensive long-term analysis, which amounts to 677,123 bike trips in total. {We scaled the bike demands by the population at the corresponding census tract and time stamp. The bike demands, like other census tract features, were linearly interpolated for the purpose of granularity.} To understand the underlying socioeconomic and demographic factors that may contribute to bike demand, we collect the corresponding features from Chattanooga during the same time frame.
\begin{itemize}
  \item \textit{Spatially, city service demand is dynamically linked to various factors.} We first generate monthly demographic and socioeconomic feature values at each bike station by associating features from 82 census tracts in Chattanooga with their closest stations. Then, we use a linear spline interpolator to obtain a finer granularity for the features, and aggregate hourly bike demand across each month for each station. For each timestamp, we calculate the correlations between the bike demand and feature data using three different correlation metrics and compute the average correlation across hours for a month. We use several correlation measures to explore different relationships between aggregated demand and features of interest. Specifically, we use Pearson's correlation coefficient, which evaluates linear relationships. 
  To measure how well monotonic functions explain the relationship between two variables, we use Spearman's rank correlation, which is a non-parametric rank statistic metric. Finally, to address the spontaneity of the data investigated, we use Kendall's rank correlation coefficient.  
  \\
  \\
  Figure \ref{fig:spatial} shows the dynamic changes of the correlation coefficients across the spatial domain. The left plot in Figure~\ref{fig:spatial} shows monthly correlations between the bike demand and the percentage of the city population aged 65 years old and over between February 2017 and January 2018. We observe the highest Pearson correlation during this time window appeared in January 2018, which amounted to -0.069, while the lowest Pearson correlation, -0.109, appeared in February 2017. The middle plot in Figure~\ref{fig:spatial} shows the correlations between bike demand and the mean household income. Unlike the left figure, we observe an overall increasing trend in the magnitudes of correlations for the middle plot. Furthermore, drastic changes in the correlation can be observed.
  \item \textit{Temporal correlations between city service demand and socioeconomic factors fluctuate significantly.} 
  
  Different from the first analysis, Figure \ref{fig:temporal} shows the temporal correlation coefficients between ridership demand and different determinants of ridership for twelve selected months from February 2017 to January 2018. For each month, the value is obtained by first aggregating bike demand data of each station during that month. Then, we calculate the correlation coefficients with respect to various features (in that month), and compute the average across all stations. More specifically, in the middle figure, we observe the temporal Pearson correlation coefficient between mean household income and bike demand decreases from 0.144 to 0.01 from February 2017 to April 2017. In the left plot of Figure~\ref{fig:temporal}, the Pearson correlation coefficient between the percentage of the city population aged 65 years old and over and bike demand shows the same trend, which decreases from 0.182 to -0.023. 
  Overall, the correlations indicate there exists unfairness in historical bike accessibility. And more importantly, the unfairness changes over time, while predicting future city states using biased historical data can lead to unfair results. 
  This observation manifests the need to adjust biased data that removes the persisting unfairness and converts the data to their unbiased counterparts with minimal changes to the existing distributions. 
\begin{figure*}[t]
\begin{minipage}[t]{.465\textwidth} 
    \includegraphics[width=\textwidth]{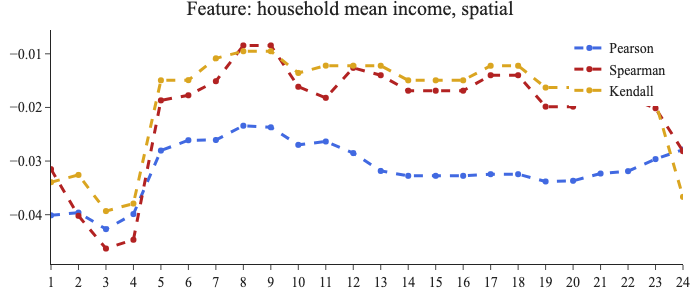}
    \caption{Spatial correlation change in 24 hours.}
    \label{fig:micro-s}
\end{minipage}%
\hfill 
\begin{minipage}[t]{.45\textwidth}
    \includegraphics[width=\textwidth]{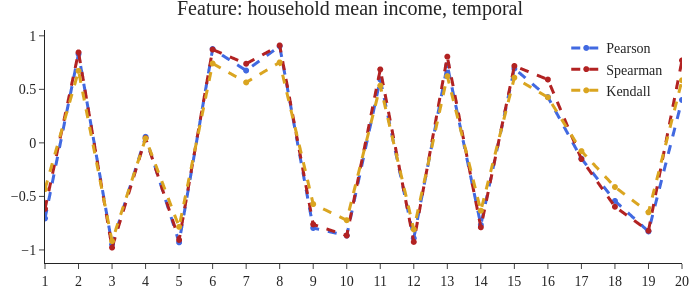}
    \caption{Temporal correlations change in 20 days.}
    \label{fig:micro-t}
\end{minipage}
\end{figure*}
  \item \textit{Micro-level dataset bias exists both temporally and spatially.} 
  Consider a data-driven model that uses historical ridership data (among other features) to predict future demand. During real-time deployment, simply analyzing the bias in monthly data can lead to unfair outcomes, as such an analysis can overlook the micro-level fluctuations, which we highlight in Figure \ref{fig:micro-s} and Figure \ref{fig:micro-t}.
  While Figure~\ref{fig:spatial} and Figure~\ref{fig:temporal} demonstrate biases in data for each month, we now analyze the inconsistency of correlations in smaller time frames. Specifically, we study hourly and daily fluctuations of the correlations between socioeconomic factors and city service demand. 
  For example, Figure \ref{fig:micro-s} shows the dynamic changes in the correlations of household mean income and bicycle demands across different stations for 24 consecutive hours. On the other hand, Figure \ref{fig:micro-t} aggregates service demand temporally and shows the dynamic change in 20 consecutive days. We observe that even on such finer scales, the relationship between the demand and socio-demographic features can change, which means that policymakers can benefit from data adjustments that act on a finer granularity. Moreover, bias exists concurrently in the temporal and spatial domains, which should be taken into account by real-time prediction models. 

\end{itemize}

\section{Overview}
\label{sec:overview}

Fairguard collects smart city states (we use \textit{states} to refer to a snapshot of the key variable of interest, e.g., ridership demand) and past socioeconomic data. The collected data, together with explicit fairness requirements provided by policymakers, are then analyzed by the Fairguard framework to produce predicted city states that follow the pre-stipulated fairness requirements. The demands provide insights into future resource allocations for city services, as shown in Figure~\ref{fig:cityOverview}.

\begin{figure}[t]
    \centering
    \includegraphics[width=.75\textwidth]{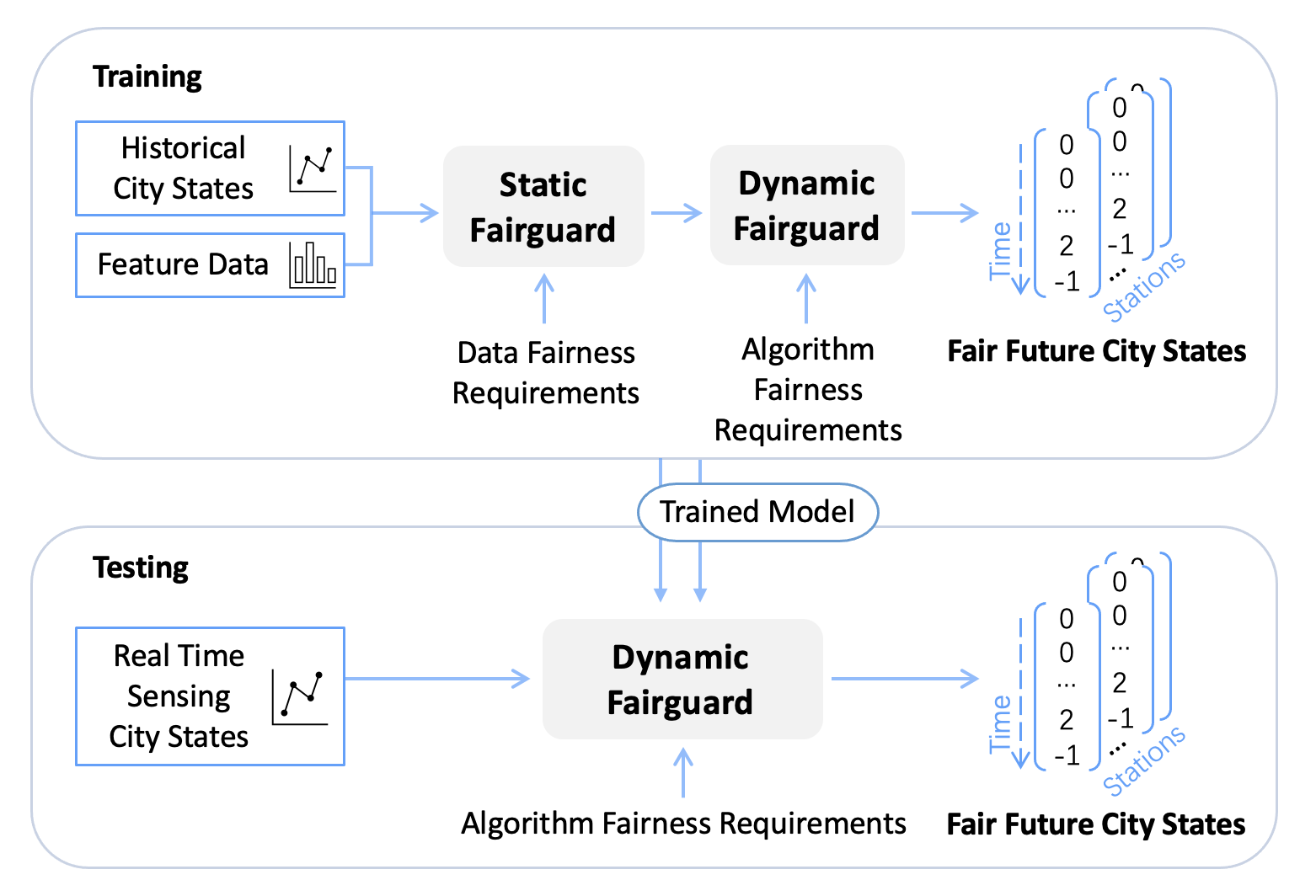}
    \caption{Training and testing phases of Fairguard}
    \label{fig:cityInternal}
\end{figure}

We show the overall training and deployment pipeline for Fairguard in Figure~\ref{fig:cityInternal}. The framework is trained offline. However, it also provides an online realization upon deployment. During the training process, we provide historical data regarding past states of a smart city, together with collected past socioeconomic data, into a static (fair) data generator called \textit{Static Fairguard}. The task of Static Fairguard is to adjust the historical states (i.e., modify the historical data) such that it follows the fairness requirements. We design Static Fairguard such that it introduces minimal changes to the distribution of the past data. The fairness-adjusted states are then fed into a dynamic algorithmic fairness adjuster called \textit{Dynamic Fairguard}. This component, via knowledge distillation, predicts future city states with the premise that the accuracy in prediction is not biased against a protected group.
Through the overall process, a trained model is produced (Dynamic Fairguard) to fairly predict future city states using unbiased past data (Static Fairguard).

When deploying the framework, real-time sensing data are directly applied to the pre-trained Dynamic Fairguard, which predicts the expected bike demands following the same fairness requirements used for training. City services can then use the prediction in fair resource allocations. 

We briefly explain the components here before diving into details. Fairness requirements are provided in Signal Temporal Logic (STL). Inside Static Fairguard, we use the modified Powell's hybrid algorithm~\cite{CHEN1981143} to construct a decoder that alters the historical states conditioned on socio-economic data based on the requirements. 
Dynamic Fairguard, on the other hand, enforces algorithmic fairness STL requirements by using a teacher-student network using the same decoder from Static Fairguard. While training, by providing fairness-adjusted data, Static Fairguard ensures that the ground truth data to Dynamic Fairguard follows the fairness requirements. Both of the components act on a high granularity of data, providing a better understanding of micro-level details.

\section{Methods}
\label{sec:methods}

\subsection{Signal Temporal Logic based Fairness Rules}

Signal temporal logic (STL)~\cite{donze2010robust} is an expressive yet rigorous formal symbolism for specifying temporal-based logic constraints. An STL formula $\varphi$ can be defined according to the following syntax \[ \varphi ::= \mu \mid \neg \varphi \mid \varphi_1 \wedge \varphi_2 \mid \varphi_1 \vee \varphi_2 \mid \square_{[a, b)} \varphi \mid \mathbin{\Diamond}_{[a, b)} \varphi \mid \varphi_1 \: \mathcal{U}_{[a,b)} \varphi_2 \] where $a, b \in \mathbb{R}_{\geq 0}$, $a \leq b$ are time steps, $\varphi_1$ and $\varphi_2$ are different STL formulae, and $\mu: \mathbb{R}^n \rightarrow \{ \top, \bot \}$ is a predicate in the form of $f(x)>0$ for a single variable $x$. Moreover, $\square$ denotes the requirement \textit{always}, $\mathbin{\Diamond}$ denotes \textit{eventually}, and $\mathcal{U}$ denotes \textit{until}. As an example, the STL specification $\, \square_{[0, 5]} \, PC_\text{SpatialMI} (X, Y) = 0 \,$ is equivalent to: \textit{the Pearson correlation between the variables $X$ and $Y$ should be zero from time 0 to 5.} Given a signal trace $\omega$, such as city states in our framework, and a set of STL formulae $\Phi$ where $ \varphi \in \Phi$, we can evaluate the satisfaction of $\omega$ with respect to $\Phi$ using the STL qualitative semantics~\cite{donze2010robust}, as established in Definition \ref{df:stl}. 
\begin{definition}[STL qualitative semantics] \label{df:stl}
Given a set of STL formulae $\varphi, \varphi_1, \varphi_2$ and a signal trace $\omega$ over a finite time domain $t \subseteq \mathbb{R}_{\geq 0}$, the qualitative semantics is defined as:

\begin{align*}
& (\omega, t) \vDash \mu & \Leftrightarrow & \quad f(x)>0 \\
& (\omega, t) \vDash \neg \varphi  & \Leftrightarrow & \quad \neg (\omega, t) \vDash \varphi \\
& (\omega, t) \vDash \varphi_1 \wedge \varphi_2 & \Leftrightarrow & \quad (\omega, t) \vDash \varphi_1 \; \text{and} \;(\omega, t) \vDash \varphi_2 \\
& (\omega, t) \vDash \Diamond_{[a, b]} \varphi & \Leftrightarrow & \quad \exists t^{\prime} \in\left[t+a, t+b\right],(\omega, t^{\prime}) \vDash \varphi \\
& (\omega, t) \vDash \square_{[a, b]} \varphi & \Leftrightarrow & \quad \forall t^{\prime} \in\left[t+a, t+b\right], (\omega, t^{\prime}) \vDash \varphi\\
& (\omega, t) \vDash \varphi_1 \mathcal{U}_{[a, b]} \varphi_2 & \Leftrightarrow & \quad \exists t^{\prime} \in\left[t+a, t+b\right], \left(\omega, t^{\prime}\right) \vDash \varphi_2 \\
&&& \quad \wedge \forall t^{\prime \prime} \in\left[t, t^{\prime}\right],\left(\omega, t^{\prime \prime}\right) \vDash \varphi_1
\end{align*}

\end{definition}

The aforementioned STL syntax and semantics serve as a template for user-specified fairness specifications. Such rules can be integrated into our proposed decoder as discussed in \ref{ssec: powell}, where fairness metrics are applied to a set of protected attributes (PA), or equivallently socioeconomic data, and city states. As an example, suppose a user provides a data fairness specification $\square_{[0, 5]} \, \allowbreak PC_\text{SpatialMI} (X, Y) = 0$. Specifically, $X$, parameterized by time $t$, represents the protected attribute data, and $Y$, parameterized by time $t$, represents the city states. The decoder returns a new set of city states $Y_\text{new}$ that satisfy the given equality, $PC_\text{SpatialMI} (X, Y_\text{new}) = 0$, in the specified time range by solving for the root of the nonlinear least-squares problem associated with the Pearson's correlation function. By Definition \ref{df:stl}, the solution $Y_\text{new}$ satisfies Equation \ref{eq:stl_example}.

\begin{equation} \label{eq:stl_example}
\begin{aligned}
    (\omega, t) \vDash \square_{[0, 5]} PC_\text{SpatialMI} (X, Y_\text{new}) = 0
    \\\Leftrightarrow \forall t^{\prime} \in\left[t, t+5\right], (\omega, t^{\prime}) \vDash PC_\text{SpatialMI} (X, Y_\text{new}) = 0
\end{aligned}
\end{equation}

On the other hand, if the user provides an STL specification with the quantifier \textit{eventually} with a time range of $[a, b]$, the time range can be instantiated to a concrete set of timestamps $[a + \alpha, b-\beta] \subseteq [a,b]$ where the specification is satisfied. The specification can then be integrated to the decoder following the method in the previous example with the \textit{always} quantifier. We provide further details in the following sections. 




\subsection{Fairness Metrics}
\label{ssec:metrics}
\subsubsection{Data Fairness Metrics}
Consider a dataset that consists of protected attributes (PA), denoted by $X$, and the values of the city states for evaluation, denoted by $Y$. For example, in the case where allocations of bikes in a city are adjusted so that the policy is uncorrelated with respect to the mean income of a neighborhood (say), $X$ corresponds to the mean income associated with each bike station, and $Y$ corresponds to the bike demand for each station. To define fairness in such contexts, we use the average fairness correlation coefficients between $X$ and $Y$. Intuitively, a larger correlation between the realization of the state variable and the protected attribute indicates a higher probability of unfairness (and vice versa).
We use Pearson’s product-moment correlation coefficient (PC) as our data fairness metric. The PC (Equation \ref{eq:1}) of variables $X$ and $Y$, according to Rodgers and Nicewander~\cite{lee1988thirteen}, is defined as the quotient of the covariance of $X$ and $Y$ and the product of the standard deviations of $X$ and $Y$.

\begin{equation} \label{eq:1}
    PC_{XY} = \frac{\sum(x_{i} - \overline{x})\sum(y_{i} - \overline{y})}{\sqrt{\sum(x_{i} - \overline{x})^2}\sqrt{\sum(y_{i} - \overline{y})^2}}
\end{equation}

In practical use cases, $X$ and $Y$ are typically arranged based on spatial or temporal orders. For cases where $X$ and $Y$ are spatially arranged, the PCs are referred to as spatial PCs. The rule applies similarly to the temporal cases.

\subsubsection{Algorithmic Fairness Metrics}



Following the same definition in prior work by Agarwal et al. \cite{agarwal2019fair}, we quantitatively evaluate model fairness using bounded group loss (BGL) as defined in Equation \ref{eq:2}, where $\zeta$ is a pre-defined threshold. 
Consider a group of PAs, denoted by $A \in \mathcal{A}$, and a regression model $\hat{Y} = f(X)$, for which $Y \in \mathcal{Y}$ is the ground truth and $X \in \mathcal{X} $ is the feature vector that can either include $A$ or not. We say the predictive model $f$ is \textit{fair} if the empirical loss $l$ satisfies Equation \ref{eq:2}. Intuitively, we take the extra step of evaluating the predictor to ensure its performance is reasonably acceptable for the protected groups. 
In our dynamic framework, we select protected groups using K-Means clustering on demographic and socioeconomic features such as household mean income. Moreover, we specify the requirements on $\zeta$ with STL and use a teacher network to supervise the model fairness. 

\begin{equation} \label{eq:2}
\mathbb{E} [\emph{l} (Y, \emph{f} (x)) | A=a]\leq\zeta \mbox{ for all }a \in A.
\end{equation}

\subsection{Fairness Adjustments}
\subsubsection{Decoder via the Modified Powell's Hybrid Algorithm}\label{ssec: powell}
One essential component of the fairness adjustment model is a method to adjust the values of the city states $Y$ with respect to a constant set of PA values, such that the STL requirements can be satisfied. This infrastructure is essential to simulate policy changes in Static Fairguard and ensure that the network in Dynamic Fairguard process enforces the STL rubrics.

We use Powell's Hybrid Algorithm~\cite{CHEN1981143} to generate the adjusted city states.
Consider a non-linear least-squares problem formulated in Equation \ref{eq: ls} below; the algorithm uses a combination of Gauss-Newton and the steepest descent method to find the root of the function $f(x)$ in Equation \ref{eq: ls}. 
\begin{equation} \label{eq: ls}
F(x) = \frac{1}{2} \lVert f(x) \rVert^2
\end{equation}

\begin{equation} \label{eq: funcl}
\underset{\delta}{min}\; L(\delta) = 2(\frac{1}{2}\epsilon^T\epsilon - (\textbf{J}^T\epsilon)^T\delta+\frac{1}{2}\delta^T\textbf{J}^T\textbf{J}\delta)\; \textrm{subject to}\;\lVert\delta\rVert\le\Delta
\end{equation}
The Powell's Hybrid method is explicitly controlled via a trust region using a quadratic model function (Equation \ref{eq: funcl}), where $\epsilon$ represents the error from the estimated measurement vector produced by the parameter function and the solution $\delta$ for the function represents the candidate step, as introduced by~\cite{1544898}, which represents the objective function only for points within a hypersphere.
Therefore, we can formulate the constrained region as a sub-problem, where $\textbf{J}$ denotes the Jacobian matrix. 

The radius of the trusted region hypersphere, $\Delta$, is typically chosen based on the success status of the previous iteration of the algorithm. In each step, the radius of the hypersphere is updated based on the success status of the previous iteration. Namely, it is based on the accuracy of the approximation to the objective function from the previous iteration.

The solution to the sub-problem can also be approximated with two trajectories: The first run from the initial coordinate to the Cauchy point is given by
\begin{equation} \label{eq: deltasd}
    \delta_{sd}=\frac{\textbf{g}^T\textbf{g}}{\textbf{g}^T\textbf{J}^T\textbf{J}\textbf{g}}\textbf{g}
\end{equation}
where $\textbf{g}$ is the steepest descent direction \begin{math} \textbf{g} = \textbf{J}^T\epsilon \end{math}. The second runs from the adjusted point to the Gaussian-Newton step \begin{math} \delta_{gn} \end{math}, where \begin{math} \textbf{J}^T\textbf{J}\delta_{gn} = \textbf{g} \end{math}, which can be solved via a perturbed Cholesky decomposition. Then, the method defines the dog leg (\begin{math}\delta_{dl}\end{math}) step: 

\begin{equation}
    \delta_{dl} =
    \begin{cases}
    \frac{\Delta}{\lVert\delta_{sd}\rVert}\delta_{sd},\, \text{if}\, \delta_{sd} > \Delta \text{ and } \delta_{gn} > \Delta
    \\
    \delta_{gn},\,\text{if}\ \delta_{gn} \le \Delta
    \\
    \delta_{new},\,\text{otherwise}
    \end{cases}
\end{equation}
where \begin{math}\delta_{new}\end{math} is the intersection between the trust boundary and the line connecting \begin{math}\delta_{sd}\end{math} and \begin{math}\delta_{gn}\end{math}. The point found in the dog leg step is then the trial point used for the next iteration.
With this method, we can find a root to a nonlinear equation (either a function related to PC or BGL in this case). Thus, we were able to design a decoder D for finding \begin{math} y_1 \end{math} such that:

\begin{equation}
    D(y_0) = y_1\;\textrm{subject to}\;S(x, \phi_i)\;\forall\; \phi_i \;\in\;T
\end{equation}

where $T$ is the set of STL requirements, $x$ is the constant feature data, \begin{math}y_0\end{math} is the preadjusted city states, \begin{math}y_1\end{math} is the decoded city states, and S is the STL converted of a given feature dataset and an STL requirement following 4.1. The adjusted city states \begin{math} y_1 \end{math} can then be used for the Static Fairguard and the Dynamic Fairguard, etc.

\subsubsection{Static Fairguard}\label{static-fair}

The purpose of the Static Fairguard is to integrate data fairness STL requirements with the decoder to statically alter the city states with respect to a corresponding PA data set, which stays constant.

Our STL requirements for the Static Fairguard is stipulated in the formula below:

\begin{equation}
\begin{aligned}
\varphi = (|PC(X_i, Y_{0_i})| >= \rho) \Rightarrow (\square_{[0,t]} (PC(X_i, Y_{1_i}) = 0) \\
\land \square_{[0,t]} (\mu_{Y_{1_i}} = \mu_{Y_{0_i}})
\land \square_{[0,t]} (\sigma_{Y_{1_i}} = \sigma_{Y_{0_i}}), 
 \forall i \in I
\end{aligned}
\label{overallSTL}
\end{equation}
where $I$ refers to a partition of the city states, $Y$, and the PA data, $X$, in either spatial or temporal dimension with a chosen granularity, with \begin{math}Y_0\end{math} being the city states before the static adjustment and \begin{math}Y_1\end{math} after adjustment. The PC requirement ensures the fairness of \begin{math}Y_{1_i}\end{math} with respect to \begin{math}X_i\end{math} for all partitions when \begin{math}Y_{0_i}\end{math} has a correlation with \begin{math}X_i\end{math} that is higher than \begin{math}\rho\end{math}, a client-defined hyperparameter. The constancy requirement on the mean and the standard deviation, together with the condition with \begin{math}\rho\end{math}, ensures minimal changes to the distribution of the original city states. For each sample in equation \ref{overallSTL}, when the if condition is not triggered, the city states will remain unchanged from the original states.

We introduce the STL requirements into the decoder constructed using modified Powell's adjustment~\cite{powell1970hybrid}, where \begin{math}D(Y_{0_i}) = Y_{1_i}\end{math} subject to the STL for all partitions. The partitions are then recombined in the same order they have been dissected to form the regulated city states, which is the output of Static Fairguard.


\begin{figure}[t]
\centering
\begin{subfigure}[b]{.6\textwidth}
  \centering
  \includegraphics[width = \textwidth]{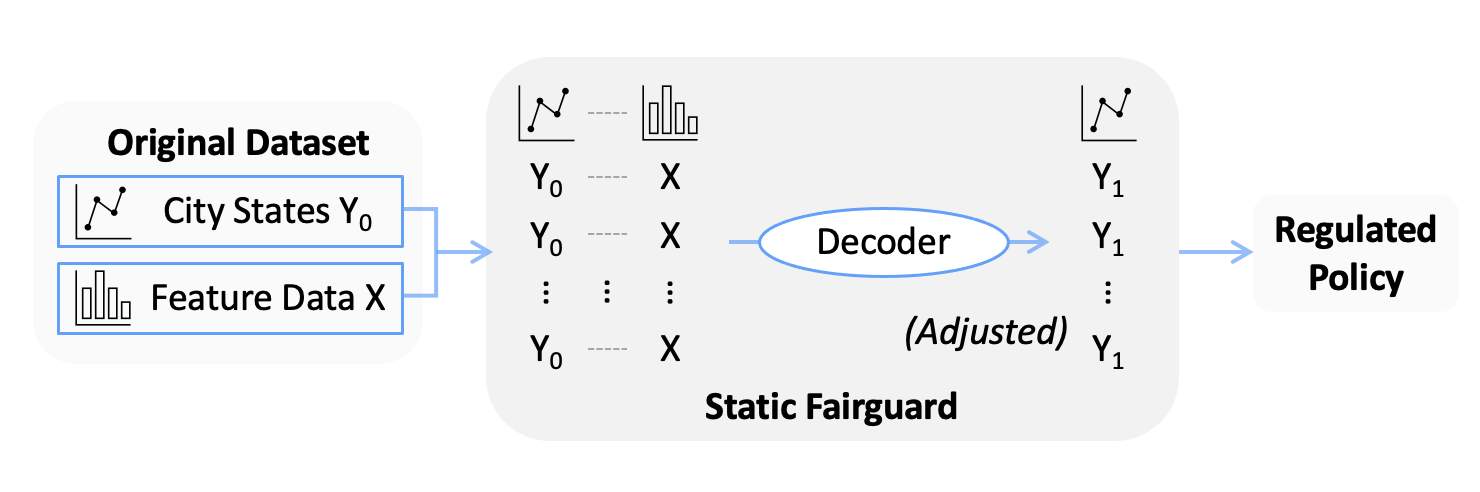}
    \caption{Static Fairguard}
    \label{fig:my_label}
\end{subfigure}%
\vspace{1em}
\begin{subfigure}[b]{.3\textwidth}
  \centering
  \includegraphics[width = \textwidth]{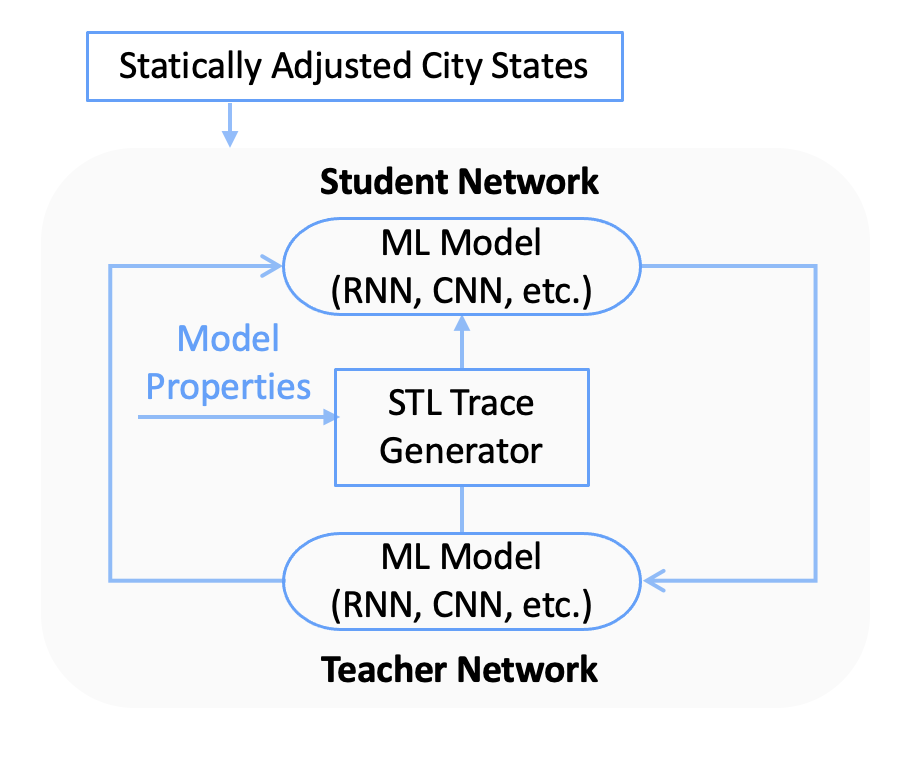}
  \caption{Dynamic Fairguard}
  \label{fig:dynamic_fair}
\end{subfigure}
\caption{Fairguard framework}
\label{fig:test}
\end{figure}

\subsubsection{Dynamic Fairguard}


The dynamic component of Fairguard is an online algorithmic fairness adjustment framework to dynamically mitigate the influence of an unfair model on the generated city states. Specifically, we consider a fair regression scenario, where the model performance on a protected group of features is evaluated quantitatively. We specify constraints on the loss of the protected group using STL, and enforce the requirement with a teacher-student network~\cite{hinton2015distilling}. We select protected groups by running a K-Means clustering on features in the dataset. For example, in our bike demand scenario, the protected group can be regional annual incomes to ensure model fairness for lower-income communities. 

\sloppy At the training phase, the multivariate student network takes in a trace of past city states (adjusted using static Fairguard), and aims to learn a predictive model $p_\theta (\boldsymbol{X})$ to generate future city states $\hat{\sigma}$ for $m$ timestamps. Let $\boldsymbol{X}^{(t)}$ denote a trace of city state at time $t$ for all features, the student network outputs $ \hat{\sigma} = (\boldsymbol{X}^{(n+1)}, \boldsymbol{X}^{(n+2)}, ... \, ,\boldsymbol{X}^{(n+m)}) = p_\theta(\boldsymbol{X}^{(0)}, \boldsymbol{X}^{(1)}, ... \, , \allowbreak \boldsymbol{X}^{(n)})$. To improve the prediction fairness for protected groups, we introduce a teacher network $q(\hat{\sigma})$ to regulate $\hat{\sigma}$. 
Let $a \in A$ denote a group of selected features. With respect to the ground truth $\sigma$, the teacher network quantitatively evaluates the BGL (Equation \ref{eq:2}) for each feature $a$ as $ l_a (\hat{\sigma}, \sigma) $, then checks $l_a$ against a set of STL requirements such as Equation \ref{eg: 1}.
\begin{equation} \label{eg: 1}
\varphi = \square_{[0,t]} (l_{a,1} \leq \zeta) \land \cdots \land \square_{[0,t]} (l_{a,n} \leq \zeta)
\end{equation}

When requirement $\varphi$ is not satisfied by prediction $\hat{\sigma}$, the teacher network applies $\varphi$ and generates a corrected prediction $\sigma'$ using the decoder, while maintaining the same mean: $\overline{\hat{\sigma}} = \overline{\sigma'}$. The student network $p_\theta$ is backpropagated with the loss function defined in Equation \ref{loss}, where $\gamma$ is a hyperparameter that regulates the strength of knowledge distillation. 
\begin{equation} \label{loss}
\mathcal{L} = \mathcal{L}(\hat{\sigma}, \sigma) + \gamma \mathcal{L}(\hat{\sigma}, \sigma')
\end{equation}

During the testing phase, using the teacher network  guarantees better requirement satisfaction, while using the student network allows for more flexibility.

\begin{table*}[t]\caption{STL rules for each case study}
\small
\centering
\begin{tabular}{|l|l|l|}
\hline
Case & Fairness Requirement & STL Formula \\
\hline
1 & 
\begin{tabular}[x]{@{}l@{}}Always adjust spatial correlation on mean income 
\\ to \textbf{$c$} if greater than \textbf{$k$}, while keeping mean and stand-
\\ ard deviation the same. 
\end{tabular}
&
$\begin{aligned}
& \varphi = (| \, PC_\text{SpatialMI}(X_i, Y_{0_i}) \, | >= k) \Rightarrow ( \square_{[0,t]} \, (PC_\text{SpatialMI} \\
& (X_i, Y_{1_i}) = c)  \land \square_{[0,t]} \, (\mu_{Y_{1_i}} = \mu_{Y_{0_i}}) \land \square_{[0,t]} \, (\sigma_{Y_{1_i}} = \sigma_{Y_{0_i}}) ) \\
&  \, \forall i \in I \\
\end{aligned}$ \\
\hline
2 & 
\begin{tabular}[x]{@{}l@{}}Always adjust spatial correlation on percentage of 
\\ households with public assistance to \textbf{$c$} if greater 
\\ than \textbf{$k$}, while keeping mean and standard deviation 
\\ the same. 
\end{tabular}
&
$\begin{aligned}
& \varphi = (| \, PC_\text{SpatialPHPA}(X_i, Y_{0_i}) \, | >= k) \Rightarrow ( \square_{[0,t]} \, (PC_\text{SpatialPHPA}  \\
& (X_i, Y_{1_i})  = c) \land \square_{[0,t]} \, (\mu_{Y_{1_i}} = \mu_{Y_{0_i}}) \land \square_{[0,t]} \, (\sigma_{Y_{1_i}} = \sigma_{Y_{0_i}}) ) \,  \\
&  \forall i \in I \\
\end{aligned}$ \\
\hline
3 & 
\begin{tabular}[x]{@{}l@{}}Always adjust temporal correlation on mean income
\\ to \textbf{$c$} if greater than \textbf{$k$}, while keeping mean and stand-
\\ ard deviation the same. 
\end{tabular} 
&
$\begin{aligned}
& \varphi = (| \, PC_\text{TemporalMI}(X_i, Y_{0_i}) \, | >= k) \Rightarrow ( \square_{[0,t]} \, (PC_\text{TemporalMI} \\
& (X_i, Y_{1_i}) = c) \land \square_{[0,t]} \, (\mu_{Y_{1_i}} = \mu_{Y_{0_i}}) \land \square_{[0,t]} \, (\sigma_{Y_{1_i}} = \sigma_{Y_{0_i}}) ) \,  \\
&  \forall i \in I \\
\end{aligned}$ \\
\hline
\multirow{2}{*}{4} & 
\begin{tabular}[x]{@{}l@{}}Always adjust spatial correlation on percentage of 
\\ household with public assistance to \textbf{$c$} if greater 
\\ than \textbf{$k$}, while keeping mean and standard deviation 
 \\ the same. 
\end{tabular}
&
$\begin{aligned}
& \varphi = (| \, PC_\text{SpatialPHPA}(X_i, Y_{0_i}) \, | >= k) \Rightarrow ( \square_{[0,t]} \, (PC_\text{SpatialPHPA} \\
& (X_i, Y_{1_i}) = c) \land \square_{[0,t]} \, (\mu_{Y_{1_i}} = \mu_{Y_{0_i}}) \land \square_{[0,t]} \, (\sigma_{Y_{1_i}} = \sigma_{Y_{0_i}}) ) \,    \\
& \forall i \in I \\
\end{aligned}$\\
\cline{2-3} & 
\begin{tabular}[x]{@{}l@{}}Always adjust spatial correlation on mean income 
\\ to \textbf{$c$} if greater than \textbf{$k$}, while keeping mean and stand-
\\ ard deviation the same. 
\end{tabular}
&
$\begin{aligned}
& \varphi = (| \, PC_\text{SpatialMI}(X_i, Y_{0_i}) \, | >= k) \Rightarrow ( \square_{[0,t]} \, (PC_\text{SpatialMI} \\
& (X_i, Y_{1_i}) = c) \land \square_{[0,t]} \, (\mu_{Y_{1_i}} = \mu_{Y_{0_i}}) \land \square_{[0,t]} \, (\sigma_{Y_{1_i}} = \sigma_{Y_{0_i}}) ) \,  \\
&  \forall i \in I \\
\end{aligned}$ \\
\hline
\end{tabular}
\label{tab:stl-case}
\end{table*}

{ \setlength{\tabcolsep}{10pt} 
\begin{table*}[t]
\centering
\small
\caption{Correlation coefficients with and without Static Fairguard}
\begin{tabular}{@{\hspace{2mm}}c@{\hspace{2mm}}|@{\hspace{2mm}}c@{\hspace{5mm}}ccc}
\toprule
\multicolumn{1}{c}{} & &  \textbf{Average PC}  & \textbf{Average SC} &  \textbf{Average KC}\\
\midrule
\textbf{Case 1} & Without Fairguard &  -0.150 & -0.124 & -0.093\\
\multirow{2}{*}{Household mean income}
& With Fairguard   &  -0.025 & -0.002 & -0.003 \\
& Improvement   &  83.333\%  &  98.387\% & 96.774\% \\
\midrule
\textbf{Case 2} & Without Fairguard &  0.1251 & 0.2247 & 0.1611\\
\multirow{2}{*}{\begin{tabular}[x]{@{}c@{}}Percentage of household \\ with public assistance
\end{tabular}} &
With Fairguard  &  0.1210 & 0.2197 & 0.1574 \\
& Improvement   &  3.277\%  &  2.225\% & 2.297\% \\
\midrule
\textbf{Case 3} & Without Fairguard & -0.0031 & -0.0022 & -0.0030\\
\multirow{2}{*}{Household mean income}
& With Fairguard   &  -0.0008 & 0.0012 & 0.0002 \\
& Improvement   &  74.194\%  &  45.455\% & 93.333\% \\
\midrule
\textbf{Case 4} & Without Fairguard & -0.150 & -0.124 & -0.093\\
\multirow{2}{*}{Household mean income} & 
With Fairguard   & -0.097 & -0.059 & -0.077\\
& Improvement  & 35.333\% &  52.419\% & 17.204\% \\
\cline{2-5}
\multirow{3}{*}{\begin{tabular}[x]{@{}c@{}} Percentage of household \\ with public assistance \end{tabular}}
& Without Fairguard & 0.1251 & 0.2247 & 0.1611\\
& With Fairguard   & 0.0890 & 0.1194 & 0.1671\\
& Improvement   & 28.857\% &  46.862\% & -3.724\% \\
\bottomrule
\end{tabular}
\end{table*}
}

\section{Evaluation}\label{sec:evaluation}

To evaluate the performance of Fairguard, we present four case studies on the Chattanooga bike sharing data and the census tract data. Case study 1 focuses on the mean household income feature with spatial fairness adjusted in Static Fairguard; case study 2 focuses on the percentage of household with public assistance feature with spatial fairness adjusted in Static Fairguard; case study 3 focuses on the MI feature with temporal fairness adjusted in Static Fairguard; case study 4 adjusts the spatial fairness in bike demands with respect to both household mean income and percentage of household with public assistance features.

For all case studies, we focus only on the last half of the dataset introduced in Section~\ref{sec:moti} because the first half of the bike demands dataset is relatively limited in changes. For all case studies, we use temporal units of hours for all components of the framework. As part of data-preprocessing, we remove all null correlations in samples of case 3. We now present methods for evaluations, as well as results regarding the  performance of the proposed approach on each test case below.

\subsection{Evaluations on Static Fairguard}

The evaluations for the Static Fairguard focus on two major performance criteria of the framework: 1) The regulated city states must obey data fairness. In other words, \begin{math}Y_1\end{math} must have a low correlation with $X$. 2) The distribution of the regulated city states must be minimally deviated from that of the original city states data to be altered. To make sure data fairness is achieved by the Static Fairguard, we measure the Pearson correlation coefficients (PC), the Spearman's rank correlation coefficients (SC), and Kendall's rank correlation coefficient (KC) of the vectors in \begin{math}Y_0\end{math} and \begin{math}Y_1\end{math} prior and after the Static Fairguard adjustment.

Apart from data fairness testing, to ensure the deviation of the adjusted city states from the original data is within an acceptable range, we also test the Mean Squared Errors (MSE) between the means of the pre-adjusted vectors in \begin{math}Y_0\end{math} and those of \begin{math}Y_1\end{math} and the MSE between the standard deviations of the same pairs of data.



When adjusting the bike demands with respect to the temporal fairness (applicable to case study 3), the temporal vectors are divided into samples with a size of 100, and adjustments are done for each of the samples for all stations. The choice of applying a dissection allows more flexibility for the clients when such frameworks are used. Better granularity controls are given to the client. At the same time, this allows less time for the decoder to process of each sample, allowing more instant feedback regarding the performances.

To allow the correct behavior of the decoder, for all case studies and all components where the decoder is used, while the cardinality of the set of requirements is smaller than the dimension of \begin{math}y_0\end{math}, zeros are appended to the decoder so that the kernel nonlinear least squares solver functions correctly. To better visualize the input rules for the Static Fairguard, we display the corresponding STL rule for each study (case 1 - case 4) in Table~\ref{tab:stl-case}, where we use $c$ to denote desired adjustment goals and $k$ to denote threshold. We specified $c$ to be $0$ and $k$ to be $0.1$ in the evaluation.

\begin{figure}[t]
    \centering
    \includegraphics[width = .75\textwidth]{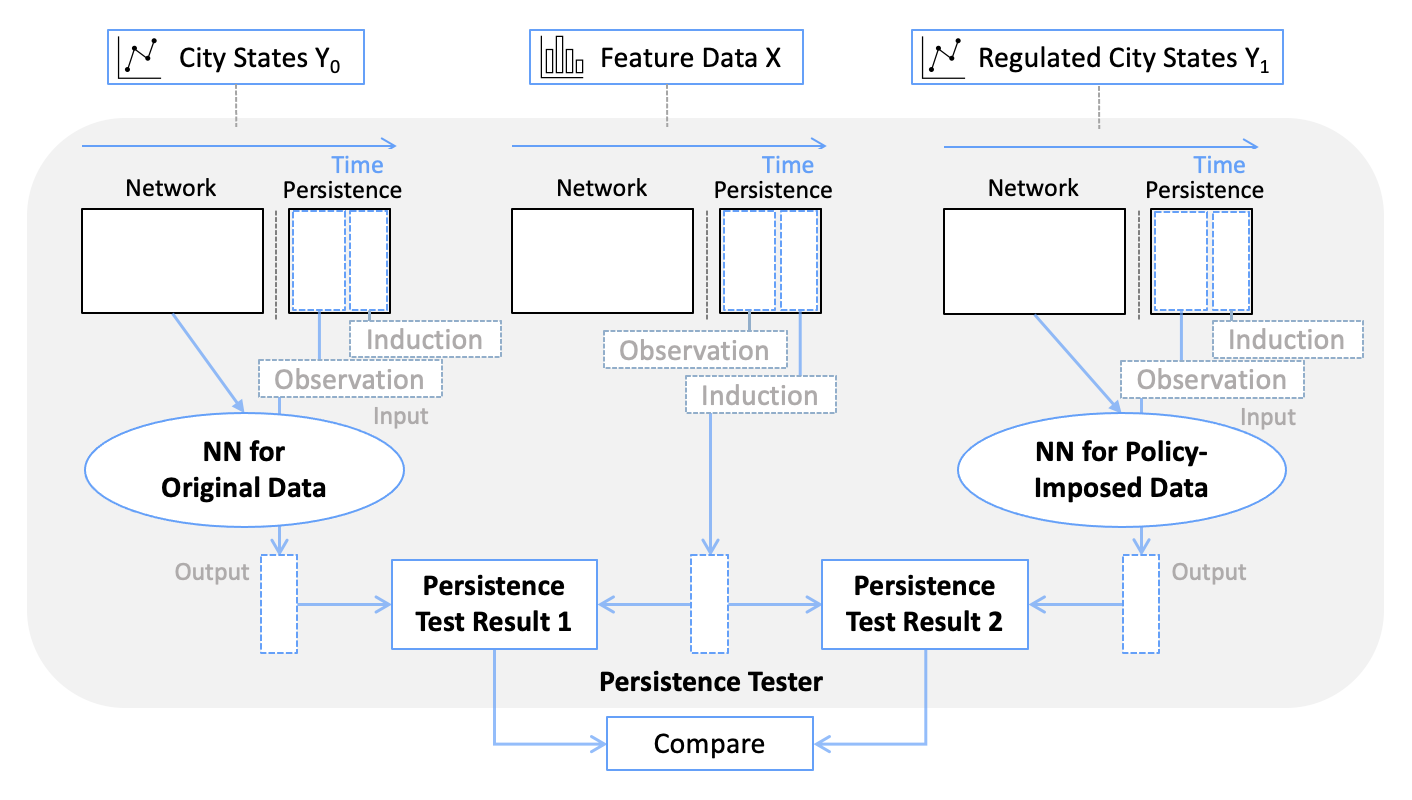}
    \caption{Persistence Testing}
    \label{fig:my_label_pt}
\end{figure}

\subsection{Persistence Testing}
One essential expectation of the framework is its long-lasting effect with stipulated requirements even if such requirements are only enforced for a short amount of time. To test such property, we propose testing Persistence Testing (PT) on the constructed framework. The input for PT is the adjusted city states of a selected PA, or equivalently the output of Static Fairguard, and PT simulates the data fairness of a given dataset after the adjusted city states are enforced for a very limited amount of time and removed.

In PT, all datasets ($X$, $Y_0$, and $Y_1$ together) were dissected into equal lengths of segments in the temporal dimension. For each of our test cases, this length is 40 units of spatial arrays. For each segment, the first 30 spatial arrays are used for the "observation phase" and the later 10 arrays are used for the "induction phase". The utilities of each phase will be discussed as we further discuss PT. Among all the segments, 80\% of randomly selected segments (let this be A) are used for constructions of two separate neural networks, and the other 20\%  (let this be B) are used for calculating the results of the persistence testing:

Specifically, to simulate the effect of the time progression with and without the adjusted city states, we use two separate neural networks (NN) that map the observation phase of each segment in A to its induction phase. With the framework being model-agnostic, for a proof of concept, we apply a convolutional neural network (CNN) architecture as each of the NN in our persistence testing. The networks \begin{math}f_0\end{math} and \begin{math}f_1\end{math} are constructed for $Y_0$ and $Y_1$ respectively: \begin{math} f_0 \end{math} and \begin{math} f_1 \end{math} are simulations of 30 days of the observation period for the Static Fairguard-unadjusted and Static Fairguard-adjusted bike demands respectively, followed by 10 days of predictions based on the corresponding observations. By doing so, we can observe the effects of only 30 days of requirement enforcement followed by its removal on data the following 10 days after its removal.
\begin{equation} \label{eq1}
\begin{split}
f_0: Y_{0_{i_\text{observation}}} \rightarrow Y_{0_{i_\text{induction}}} \\
f_1: Y_{1_{i_\text{observation}}} \rightarrow Y_{1_{i_\text{induction}}}
\end{split}
\end{equation}



To construct the network, for each test case, we allocated 80\% of data for training, 15\% for testing, and 5\% for validation. The fully trained network is then applied to observation phases of one randomly selected sample of persistence data to produce predictions of \begin{math}Y_0\end{math} and \begin{math}Y_1\end{math} data with the length of the induction phase. The PCs of the predictions with respect to the corresponding intact PA data were calculated and compared. If the requirement stipulated using the Static Fairguard carries out effects into the future (with only limited enforcement time), the PCs for the regulated city states are expected to be closer to 0. The persistence testing is done 5 times each for the case studies to account for the randomness in the testing process.

\subsection{Evaluations on Dynamic Fairguard}
We evaluate the Dynamic Fairguard framework using four cases of Chattanooga bike sharing data with different socioeconomic and demographic features. 
For each evaluation case, Dynamic Fairguard is used in conjunction with different predictive models that take in real-time past bike demand and output fair future bike demand. 
For all Dynamic Fairguard evaluation scenarios, we consider 42 bike sharing stations in total, and select protected groups using K-Means on their socioeconomic or demographic features. 
More specifically, for cases 1 and 3, six stations are selected as the protected stations, in case 2, four stations are selected, while in case 4, seven stations are set as the protected groups. 
Additionally, the models used in case 1 are trained on fair data adjusted by Static Fairguard considering spatial correlations of bike demand and household mean income, while the models used in case 2 are trained on Fairguard-adjusted data considering spatial correlations of bike demand and the percentage of households with public assistance. For the models used in case 3, we use Static Fairguard to pre-adjust the training data considering temporal correlations between household mean income and regional bike demand. For case 4, we take into account two features spatially, household mean income and percentage of households with public assistance, at the same time.

\begin{table*}[t]
\small
\centering
\caption{Persistence Testing results}
\begin{tabular}{|c|ccc|ccc|ccc|}
\hline
\multirow{2}{*}{Trial} & \multicolumn{3}{c|}{\textbf{Case 1: Household mean income}} & \multicolumn{3}{c|}{\textbf{Case 2: Percentage of public assistance}} & \multicolumn{3}{c|}{\textbf{Case 3: Household mean income}} \\
                       & Unadjusted      & Fairguard      & Improvement     & Unadjusted         & Fairguard         & Improvement         & Unadjusted      & Fairguard      & Improvement     \\ \hline
1                      & -0.096          & -0.003         & 96.875\%        & 0.1135             & 0.1149            & -1.2335\%           & -0.0155         & -0.0069        & 55.484\%        \\
2                      & -0.091          & -0.038         & 58.242\%        & 0.1166             & 0.1156            & 0.8576\%            & -0.1100         & 0.0045         & 95.909\%        \\
3                      & -0.206          & -0.003         & 98.544\%        & 0.1272             & 0.1237            & 2.7516\%            & -0.0385         & 0.0311         & 19.221\%        \\
4                      & -0.086          & -0.073         & 15.116\%        & 0.1193             & 0.1150            & 3.6044\%            & 0.0367          & -0.0315        & 14.169\%        \\
5                      & -0.191          & -0.004         & 97.906\%        & 0.1454             & 0.1415            & 2.6823\%            & 0.1389          & -0.0044        & 96.832\%        \\ \hline
\end{tabular}
\end{table*}

\begin{table*}[t]
\small
\centering
\begin{tabular}{|c|ccc|ccc|}
\hline
\multirow{2}{*}{Trial} & \multicolumn{3}{c}{\textbf{Case 4: Household mean income}} & \multicolumn{3}{c|}{\textbf{Percentage of public assistance}} \\
                       & Unadjusted      & Fairguard      & Improvement     & Unadjusted         & Fairguard         & Improvement         \\ \hline
1                      & 0.1384          & 0.0128         & 90.751\%        & -0.211             & -0.023            & 89.100\%            \\
2                      & 0.1092          & -0.0161        & 85.256\%        & -0.198             & 0.005             & 97.475\%            \\
3                      & 0.1144          & 0.1036         & 9.441\%         & -0.082             & -0.085            & -3.659\%            \\
4                      & 0.1486          & 0.0055         & 96.299\%        & -0.211             & 0.001             & 99.526\%            \\
5                      & 0.1368          & -0.0270        & 80.263\%        & -0.216             & 0.057             & 73.611\%            \\ \hline
\end{tabular}
\end{table*}

As a preprocessing step, we normalize all data before training. Moreover, we consider the following metrics in the evaluation for the dynamic component. Firstly, we calculate the overall mean squared error (MSE) for all groups and compare the accuracy of the predictive models with and without Fairguard. 
Secondly, we evaluate MSE (PA), the average MSE for protected groups, to compare how much Fairguard can help boost model performance on them. 
Thirdly, we report BGL (\%), which is the percentage of protected groups with MSE lower than the overall MSE. Table \ref{tab:eval-dyn} shows the evaluation results of the Dynamic Fairguard. 

{ \setlength{\tabcolsep}{10pt} 
\begin{table*}[htp]
\small
\centering
\caption{Model performance with and without Dynamic Fairguard}\label{tab:eval-dyn}
\begin{tabular}{@{\hspace{2mm}}c@{\hspace{2mm}}|*{7}{@{\hspace{2mm}}c@{\hspace{2mm}}}}
\toprule
& \textbf{Model} & \textbf{MSE} & \textbf{MSE (PA)} &  \textbf{BGL (\%)} & \specialcell{ \textbf{Fairguard} \\ \textbf{MSE}} & \specialcell{ \textbf{Fairguard} \\ \textbf{MSE (PA)}} & \specialcell{ \textbf{Fairguard} \\ \textbf{BGL (\%)}} \\
\midrule
\textbf{Case 1} & LSTM & 0.6054 & 0.5542 & 85.71 & 0.6279 & 0.5081 & 100.00\\
\multirow{3}{*}{\begin{tabular}[x]{@{}c@{}}Household \\ mean income
\end{tabular}} & ConvLSTM &  0.7688 & 0.8719 & 14.29 & 0.7491 & 0.8806 & 14.29\\
& CNN-LSTM &  0.9705  & 0.7914 & 50.00 & 0.9844 & 0.7731 & 50.00\\
& DCRNN & 0.1859 & 0.1759 & 51.43 & 0.1870 & 0.1728 & 57.14 \\

\midrule

\textbf{Case 2} & LSTM & 0.6667 & 0.6270 & 75.00 & 0.6622 & 0.6329 & 75.00\\
\multirow{3}{*}{\begin{tabular}[x]{@{}c@{}}Percentage of \\ public assistance
\end{tabular}} & ConvLSTM &  0.7767 & 0.4544 & 37.50 & 0.7992 & 0.4366 & 37.50\\
& CNN-LSTM &  0.8724  & 0.6227 & 50.00 & 0.8593 & 0.5954 & 50.00\\
& DCRNN & 0.2522 & 0.2525 & 40.00 & 0.2167 & 0.2378 & 55.00\\

\midrule

\textbf{Case 3} & LSTM & 0.6249& 0.6377 & 57.14 & 0.6497 & 0.6304 & 42.86\\
\multirow{3}{*}{\begin{tabular}[x]{@{}c@{}}Household \\ mean income
\end{tabular}} & ConvLSTM &  0.7975 & 0.8454 & 35.71 & 0.7790 & 0.8542 & 28.57\\
& CNN-LSTM &  0.8551  & 0.8944 & 35.71 & 0.8818 & 0.9045 & 35.71\\
& DCRNN & 0.2359 & 0.2461 & 57.14 & 0.2389 & 0.2441 & 60.00\\

\midrule

\textbf{Case 4} & LSTM & 0.6851& 0.6788 & 66.67 & 0.6942 & 0.6582 & 50.00\\
\multirow{3}{*}{\begin{tabular}[x]{@{}c@{}}Household mean income \\ and percentage \\ of public assistance
\end{tabular}} & ConvLSTM &  0.7796 & 0.9424 & 33.33 & 0.8066 & 0.9900 & 25.00\\
& CNN-LSTM &  0.8536  & 0.9976 & 41.67 & 0.8654 & 0.9963 & 41.67\\
& DCRNN & 0.2277 & 0.2357 & 50.00 & 0.2050 & 0.2170 & 56.67\\
\bottomrule
\end{tabular}

\end{table*}
}

\subsection{Results Summary and Reflections}

\subsubsection{Results for Case study 1}
 For Case 1, 26,285 samples are adjusted (due to the condition on \begin{math}\rho\end{math}) out of 37,596 samples in Static Fairguard. The adjustment coverage is 69.914\%. The Mean Squared Error between the means of the original states of the city and the Static Fairguard-adjusted states is $ 4.443 \times 10^{-31} $, and the MSE on the standard deviation is $ 5.237 \times 10^{-19} $. 

Observing the results, we notice an improvement of over 80\% for all three metrics for the performances of the static component of Fairguard, with the improvement on KC and SC close to 100\%. Noticeably, the changes in Spearman's and Kendall's rank correlation coefficients, which are not explicitly stipulated as targets of adjustments by the decoder, outperform that of the Pearson correlation coefficient. This suggests that the Pearson correlation coefficient acts as a comprehensive requirement to the decoder, which does not limit the ability of the Static Fairguard to the only data-fairness rule stipulations. The trivial deviations in the means and standard deviations of the samples before and after the Static Fairguard adjustment also reflect the capacity of the decoder in preserving the pre-adjustment distribution. It also confirms the choice of applying the decoder only if the existing correlation violates the minimum threshold hyperparameter. All samples for persistence testing show improvements in the static component. This testifies to its ability to perform fine-granularity adjustments considering the limited length of intervals in Persistence Testing.

For Dynamic Fairguard performances, with an overall decrease in the prediction loss of the protected attribute, we also see a slight increase in the overall loss. The increase in the percentage of BGL suggests the tradeoffs between the prediction accuracies and the algorithmic fairness. It also reflects the capacity of the framework in improving the prediction model fairness. Among all the models selected for the dynamic component, diffusion convolutional recurrent neural network (DCRNN) performs the best in terms of prediction. Both LSTM and DCRNN kernels manifest an increase in the percentage of BGL, with LSTM achieving 100\%.

\subsubsection{Results for Case study 2}
For Case 2, a total of 36,519 samples are adjusted (due to the condition on \begin{math}\rho\end{math}) out of 37,596 samples in Static Fairguard. The adjustment coverage is 97.135\%. The MSE between the means of the original states of the city and the Static Fairguard-adjusted states is $2.390 \times 10^{-8}$, and the MSE on the standard deviations is $9.975 \times 10^{-7}$. 

Unlike mean income, which has a high range across time, the percentage of household with public assistance is only valid between 0 and 100. This suggests a lower standard deviation, and thus the capacity of the decoder in the static component is limited by the STL requirement on standard deviations. Therefore, we notice a significantly smaller improvement by the decoder in the Static Fairguard. However, general improvements are still present in both the Static Fairguard performance and the Persistence Testing results, suggesting that the Fairguard consistently meets the expectations of data fairness. The same pattern of decreasing loss on the predictions of protected attributes still persists in Case 2, yet the difference between the general prediction loss before and after the Dynamic Fairguard is trivial for all models throughout this case study.

\subsubsection{Results for Case study 3}
For Case 3, the MSE between the means of the original states of the city and the Static Fairguard-adjusted states is 0.1109, and the MSE on the standard deviation is 1.1279. 


The deviations for the means as well as the standard deviations between the Fairguard unadjusted and adjusted states are larger in this case study. This is because for consistency with other case studies, the measurements on distribution statistics are taken across the entire dataset. However, the STL requirements are stipulated on every 100 timestamps of each station, as suggested in Section 5.1. Apart from this, both the performance testing and the Static Fairguard performance results show significant improvements in data fairness, with a decrease of 93.333\% in the magnitude of the average Kendall's rank correlation coefficients. In terms of evaluations on the dynamic component, 2 out of 4 tested kernels show a decrease in the prediction loss of the protected attribute.

\subsubsection{Results for Case study 4}
For Case 4, the MSE between the means of the original states of the city and the Static Fairguard-adjusted states is $3.768 \times 10^{-7}$, and the MSE on the standard deviation is $1.366 \times 10^{-5}$. 



Results in the static component indicate a higher improvement in data fairness in the percentage of households with public assistance. This is caused by the correlation between this selected feature and household mean income, both of which are used in the STL requirements for data fairness. In terms of the dynamic component, among all cases, DCRNN consistently outperforms other models in prediction accuracy and ensures BGL increase in all cases. It is thus considered the most suited kernel for the Dynamic Fairguard. Overall, in this case study, the magnitudes of improvements with respect to both of the selected socioeconomic features indicate the potential of the Fairguard for working with multiple fairness stipulations simultaneously.

\section{Related Work}
\label{sec:relatedwork}

\paragraph{Fairness metrics} The definition of fairness has been a much debated topic in many fields, such as social science, public health, and political science. Machine learning frameworks, however, can be evaluated with numerous quantitative fairness metrics that focus on different aspects. One important aspect of a fair model is to ensure that model decisions are not based on protected characteristics. For example, in a fair classification scenario, a desired constraint is that the model does not classify based on predefined attributes such as income level and gender. A quantitative metric for this requirement is demographic parity (DP), which constrains the statistical correlations between the predictions and protected attributes. Different from DP, equalized odds (EQ) constrains the conditional independence of predictions and protected attributes~\cite{mehrabi2021survey}. In contrast, another work~\cite{kearns2019empirical} has highlighted the importance of ensuring a good model performance on protected subgroups, which is also known as subgroup fairness. In subgroup fairness, it is acceptable for the overall performance to be sub-optimal in order to favor the pre-defined subgroups. Bounded group loss (BGL), as an example, requires the model to calculate the loss for protected groups during training and testing time, and claims that the model is fair when the loss of protected groups is lower than a constraint value. Another example is treatment equality which requires the same ratio of false negatives and false positives across all protected groups. In this work, we highlight BGL as a model fairness measurement aided with Pearson’s product-moment correlation coefficient as data fairness measurement. 

\paragraph{Fair machine learning}
Although there is not a single fairness definition for all models, many methods have been developed by the research community to foster fairness in a machine learning process. Specifically, the bias of a ML-based framework can come from the training data, algorithm design, and user interaction. Therefore, various pre-processing, in-processing, and post-processing methods are developed and can be used in conjunction with each other~\cite{ma2021predictive}. 
Pre-processing methods that modify training data or generate new synthetic data with generative models have been proposed to remove bias. However, those methods lack formal guarantees on the data properties. To mitigate this, we use a logic-based approach in this work to correct the original data while allowing the new data to satisfy important properties. Developing an in-processing method is usually tied to the fairness metrics and type of model. For example, ~\cite{zafar2017fairness} proposes a fair classifier that maximizes fairness with accuracy constraints and~\cite{chzhen2020fair} develops a fair regression model evaluated with Demographic Parity and formulas the objective as solving for the Wasserstein barycenter. However, our approach is different from them as we integrate formal logic to constrain the data and model fairness. Our approach to exploring the benefits of STL in learning algorithms is inspired by previous work, STLnet~\cite{ma2020stlnet}. The proposed network accepts pre-defined specifications and uses a teacher-student network to ensure the rule satisfaction rate. Our framework extends STLnet for fairness concerns in the dynamic generator, and marries fairness metrics to STL specifications. Moreover, our dynamic generator is model-agnostic and adapts well to predictive tools such as recurrent networks, convolutional networks, and graph neural networks. 

\paragraph{Fair smart city designs}
Overall, situating fairness definitions and methods into context is necessary when applying these methods. For example, the goal of training a fair diagnostic classifier can be different from training a fair language model. In smart city designs, fairness concerns are focused on resource allocation, energy harvesting, social equity, and so on~\cite{ma2018cityresolver, ma2021toward}. Focusing on 311 requests in Kansas City, MO, the authors~\cite{KONTOKOSTA2021102503} study the fairness in city datasets and designs a model to identify bias. Our work goes beyond just classifying the lack of fairness by also taking actions to mitigate the effect and making sure the system is not biased against the protected communities.~\cite{9216899} considers fair traffic light control for smart cities to minimize the average vehicle waiting time at intersections, which uses the fairness criterion as the reward in a reinforcement learning algorithm. However, in real smart city applications with multiple biased influences, designing a good reward can be challenging. Instead, our work uses logic-based fairness specification as fairness constraints, which is much more intuitive, adaptive, and expressive. 

\section{Discussions}

As a complex optimization problem, resource allocations in smart cities require tradeoffs, and future directions for the framework include solutions in controlling the scales of such tradeoffs and adding more flexibility in the range of requirements acceptable by the framework. 

\paragraph{Tradeoff Control} With an unaltered kernel choice for both the baseline models and the Dynamic Components, improvements in the overall prediction accuracy cannot be readily achieved. Thus, to ensure fairness in the prediction losses of the protected attribute with respect to the total population, one can expect an increase in the accuracy for the protected group and a decrease in the overall performance. Furthermore, another tradeoff exists in the Static Fairguard. Namely, higher data fairness requires more changes to existing city states, generating more deviations from the original city states in the Fairguard processed outcome. Future research can focus on allowing clients to determine the weight of each requirement, allowing a higher degree of control and a better comprehension of the STL requirements by the Fairguard.

\paragraph{Flexibility in requirements} In this paper, we built a framework to enhance fairness requirements in deep learning-based prediction models. To be noted, as a system designer, we do not specify the requirements. Instead, the requirements are provided by the city policymakers. There are existing tools~\cite{chen2022cityspec} for a non-expert in formal methods to specify requirements in English and convert them into formal specifications automatically. Our system is general to incorporate the requirements provided by the users. 

Furthermore, the decoder is essentially a root-solver for both linear and nonlinear functions, generated using STL requirements. Therefore, the Static Component can be extended to solving other resource allocation problems outside of the scope of fairness control. Such applications include the prediction of future bike demands with the closing of a station and the opening of a new station, etc. However, not all STL requirements are compatible with the decoder constructed using modified Powell's Hybrid Methods. The Spearman's and Kendall's rank correlation coefficients, for instance, as categories of rank correlations, cannot be decoded by the current Static Fairguard. When new rules are given to the Fairguard, the differentiability of converted functions is crucial to the success of the decoder results. Alternatively, other numerical approaches in root finding can be studied as kernels to the decoder in the Fairguard. The decoder will also benefit from solutions to integer programs as opposed to their relaxations when an approximation to the nearest integers is not sufficient in some resource allocation problems in a smart city, especially when the objective functions are linear and integer constraints are given. 


\section{Conclusion}
\label{sec:conclusion}

In this paper, we present Fairguard, an STL-based resource allocation framework for data fairness of smart city states with respect to socioeconomic data and algorithmic fairness of prediction models. The framework consists of a static component using modified Powell's hybrid methods for data fairness adjustment and a dynamic component using a teacher-student network for ensuring algorithmic fairness in predictions. Four comprehensive case studies, focusing on both temporal and spatial fairness, are conducted on the bike demands dataset from Chattanooga, TN. Fairguard achieves over 80\% improvement in data fairness in all metrics of case 1. In terms of algorithmic fairness, the LSTM-based model in the dynamic component of Fairguard in case 1 achieves Bounded Group Loss 100\% of samples tested, and the DCRNN-based model, when incorporated with the Fairguard, achieves Mean Squared Errors of less than 0.25 in all cases of predictions for both the overall performances and the performances on the protected attributes.

\bibliographystyle{unsrt}
\bibliography{sample-base}

\end{document}